\documentclass{article}

\usepackage[preprint]{neurips_2026}

\usepackage[utf8]{inputenc} 
\usepackage[T1]{fontenc}    
\usepackage{hyperref}       
\usepackage{url}            
\usepackage{booktabs}       
\usepackage{amsfonts}       
\usepackage{nicefrac}       
\usepackage{microtype}      
\usepackage{xcolor}         
\usepackage{amsmath}
\usepackage{adjustbox}
\usepackage{tikz}
\usepackage{amsmath} 
\usepackage{amssymb} 
\usepackage{xcolor}  
\usepackage{caption}
\usepackage[ruled,vlined]{algorithm2e}
\usepackage{multirow}
\usepackage[utf8]{inputenc}
\usepackage{graphicx}
\usepackage{caption}
\usepackage{multirow}
\usepackage{booktabs}
\newcommand{\mask}{\texttt{[MASK]}}
\newcommand{\drean}{\texttt{Dream-v0-Instruct-7B}}
\newcommand{\llada}{\texttt{ LLaDA-1.5 }}
\newcommand{\lladanew}{\texttt{LLaDA-8B-Instruct}}

\usetikzlibrary{arrows.meta, positioning, shapes.geometric, calc, backgrounds, fit, matrix, shadows}

\title{Mean-Field Parallel Decoding for\\ Discrete Diffusion Language Models}

\author{%
  Tamim Zoabi \\
School of Electrical \& Computer Engineering \\    
Tel Aviv University, Israel \\
\texttt{tamimzoabi@mail.tau.ac.il} \\
  \And
  Ameen Ali \\
  School of Computer Science and AI \\
  Tel Aviv University, Israel \\
\texttt{ameenali@mail.tau.ac.il} \\
  \AND
  Liran Ringel \\ Department of Computer Science \\
Technion, Israel Institute of Technology, Israel\\
\texttt{liranringel@cs.technion.ac.il} \\
  \And
  Lior Wolf \\
  School of Computer Science and AI \\
  Tel Aviv University, Israel \\
  \texttt{wolf@cs.tau.ac.il}
}

\begin{document}

\maketitle

\begin{abstract}
Discrete diffusion language models enable parallel token generation, offering a pathway to low-latency decoding. However, selecting tokens independently by marginal confidence limits effective parallelism: tokens that appear reliable in isolation can form incompatible configurations when several positions are updated at once.
We introduce a training-free decoding framework that coordinates these parallel updates. At each forward pass, the method assigns a commit score to each masked position and refines these scores using pairwise interactions derived from the model's predictive distributions. A variational relaxation yields a simple fixed-point update that suppresses conflicting simultaneous commitments within a single forward pass. This mechanism allows the decoder to commit more tokens in parallel while maintaining competitive generation quality. The method is lightweight, requires no auxiliary model or retraining, and drops into existing diffusion decoding pipelines without modification. Experiments on reasoning and code-generation benchmarks show consistent improvements in the quality-latency trade-off.

Code: \href{https://github.com/AmeenAli/MDP}{GitHub}
\end{abstract}

\section{Introduction}
\label{sec:intro}

Large-scale language models~\citep{achiam2023gpt,bai2023qwen,comanici2025gemini,liu2024deepseek} have traditionally relied on autoregressive (AR) decoding, which generates tokens sequentially from left to right. While AR models achieve state-of-the-art performance across a wide range of tasks, their sequential nature imposes a fundamental bottleneck on inference latency, as the generation of each token requires a full forward pass of the model. To overcome this limitation, non-autoregressive  and semi-autoregressive models have emerged, aiming to generate multiple tokens in parallel to improve throughput and reduce wall-clock time.

Among these approaches, Discrete Diffusion Language Models (dLLMs) \citep{austin2021structured, li2022diffusionlm,zhu2025llada,nie2026large,bie2026llada21speedingtextdiffusion} have gained significant attention. By treating text generation as an iterative denoising process, dLLMs allow for flexible decoding schedules where multiple masked positions can be unmasked simultaneously. However, aggressive parallelism in dLLMs introduces a critical challenge: ``joint inconsistency''~\citep{ringel2026dependency,dawn,zhang2026generation}. Most existing dLLMs employ a factorized reverse policy that predicts token distributions for each masked token independently~\citep{ye2025dream,zhu2025llada}. While this independence assumption simplifies the model architecture, it fails to capture the complex linguistic dependencies, such as syntactic agreement and semantic coherence that exist between tokens. Consequently, when multiple tokens are committed in parallel based solely on their marginal confidences, the resulting sequence often suffers from ``joint mismatch,'' where tokens are individually plausible but mutually incompatible.

Several recent approaches attempt to reduce this inconsistency by modifying either the decoding procedure or the underlying diffusion formulation. Dependency-aware inference methods such as DAWN~\citep{dawn} explicitly estimate inter-token dependencies and use them to schedule safer parallel commitments, while related analyses of diffusion decoding show that marginally confident tokens can still conflict when committed jointly~\citep{ringel2026dependency}. Another line of works, including semi-autoregressive and block diffusion models, constrains the generation process by decoding autoregressively at the block level while refining tokens within each block in parallel~\citep{bie2025llada20scalingdiffusionlanguage, bie2026llada21speedingtextdiffusion}. This blockwise formulation improves stability relative to fully unconstrained parallel decoding, but it does not by itself resolve the dependency structure among tokens committed within the same block. Other strategies rely on auxiliary verification, repeated influence estimation, or training objectives that encourage stronger joint modeling; however, these methods can introduce additional computation, require extra model components, or depend on retraining the base model. These limitations motivate a lightweight, training-free mechanism that estimates token interactions during inference and uses them to regulate which positions can be safely committed together.

\begin{figure}[t]

    \centering
     \hfill
    \includegraphics[width=1\linewidth,height=0.4\textwidth]{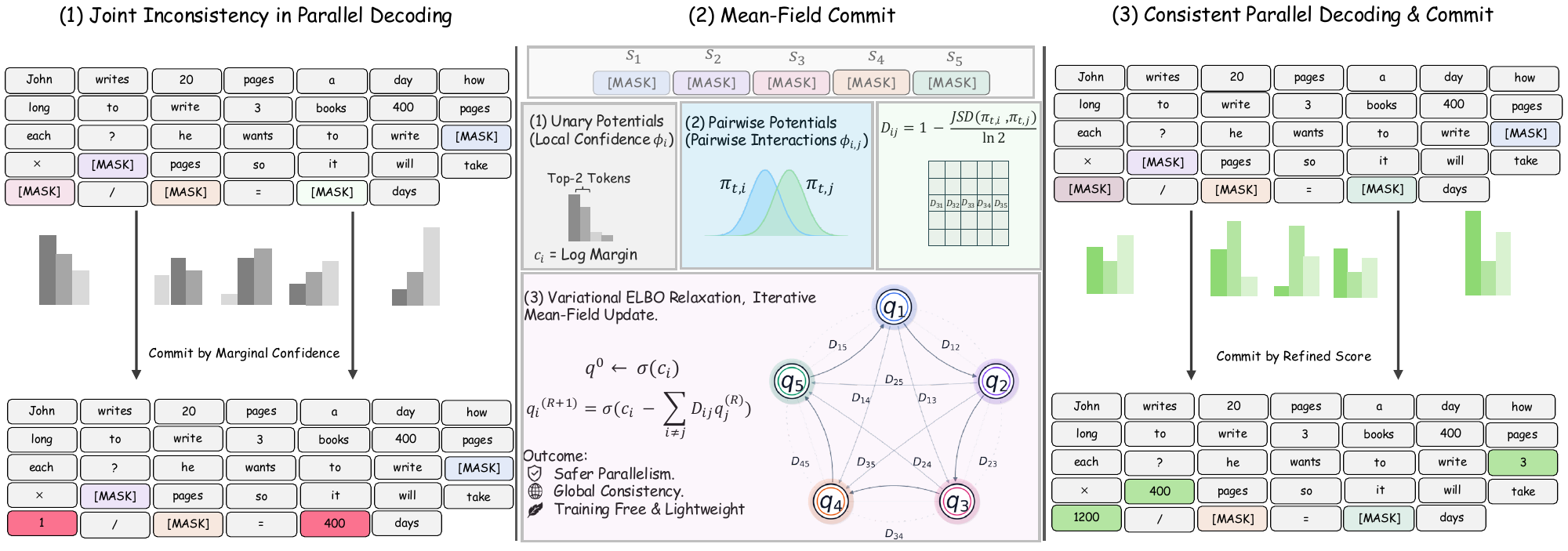}
     \hfill \vspace{-2mm}
    \caption{\textbf{Mean-Field Parallel Decoding Framework.} (Left) Joint Inconsistency: Standard parallel decoding commits tokens based on position-wise confidence (Top-3 confidence shown). This ignores inter-token dependencies, leading to an incompatible sequence (e.g., repeating the token ``3''). (Center) Mean-Field Commit: Our framework frames joint token commitment as a variational mean-field optimization. It computes local margins ($c_i$) and pairwise similarities ($D_{ij}$ using Jensen-Shannon Divergence), then iterates (R=2) to update commit intensities $q_i$. (Right) Consistent Parallel Decoding \& Commit: The refined, global-aware selection enables consistent parallel updates. This induces inhibitory signals that unmask only non-conflicting high-confidence positions, yielding a more reliable parallel update while increasing parallelism.}
    \label{fig:mainfig}
\end{figure}

In this paper, we present a novel, single-pass decoding framework: \textbf{Mean-Field Parallel Decoding}. As illustrated in Figure~\ref{fig:mainfig}, we formulate parallel token commitment as a structured inference problem over binary variables, moving away from the ``joint inconsistency'' of marginal-based selection (Figure~\ref{fig:mainfig}, left). We construct pairwise interactions defined on masked positions directly in the model’s output space using the Jensen-Shannon divergence (JSD) between predictive distributions. This yields a stable signal of agreement and competition between candidate tokens (Figure~\ref{fig:mainfig}, center). By deriving a variational mean-field relaxation, we obtain a fixed-point update in which high-confidence positions modulate others through these interactions. This induces interaction-aware commit decisions, enabling increased parallelism without sacrificing coherence (Figure~\ref{fig:mainfig}, right), all within a single forward pass and without auxiliary models or retraining.

Our contributions are summarized as follows: 

(1) We formalize the parallel commit selection problem in dLLMs as a structured inference task, providing a theoretical bridge between iterative denoising and energy-based models. (2) We introduce a training-free, gradient-free metric based on distributional overlap (JSD) that requires only a single forward pass and no auxiliary models. (3) We derive an efficient mean-field iteration that converges rapidly to a stable commit set, allowing for adaptive parallelism that scales with the model's local certainty and interaction-aware commitment and (4) We demonstrate through extensive experiments that our method significantly improves the quality-latency trade-off in masked diffusion decoding, achieving superior generation coherence while delivering the highest throughput among all compared methods, with minimal computational overhead.

\section{Related Work}

Discrete diffusion language models (dLLMs) extend denoising diffusion to discrete token spaces, enabling iterative refinement and parallel decoding~\citep{austin2021structured, li2022diffusionlm, hoogeboom2021argmax}. Recent large-scale instantiations (e.g., LLaDA, Dream)~\citep{ye2025dream,nie2026large,bie2025llada20scalingdiffusionlanguage,bie2026llada21speedingtextdiffusion} demonstrate competitive performance with flexible decoding schedules while maintaining a factorized reverse parameterization over token positions.

A central direction focuses on accelerating generation via parallel token commitment. In autoregressive models, speculative decoding and its extensions~\citep{leviathan2023fast, chen2023accelerating, kim2025speculative, chen2026dflash,ringel2026ddtree} generate multiple tokens and verify them with a target model. In diffusion-based decoding, token selection is typically guided by local uncertainty metrics such as entropy or margin~\citep{ghazvininejad2019mask,ye2025dream,nie2026large}, as well as more recent methods including KL-based selection KLASS ~\citep{kim2026klass} that uses KL-based selection, LocalLeap~\citep{kong2025accelerating}, which improve throughput by prioritizing high-confidence positions. These approaches rely on marginal statistics and do not explicitly account for dependencies between simultaneously committed tokens.

Modeling inter-token dependencies in non-autoregressive generation has been explored through iterative refinement~\citep{ghazvininejad2019mask}, latent-variable formulations~\citep{gu2018nonautoregressive}. In the context of diffusion language models, recent work has begun to explicitly address the limitations of factorized decoding. Adaptive parallel decoding (APD)~\citep{israel2026accelerating} incorporates an auxiliary autoregressive model to approximate joint effects. More recent works directly model dependencies at decoding time: DAPD~\citep{kim2026dependency} constructs an attention-induced dependency graph and selects independent token subsets, DAWN~\citep{dawn} estimates inter-token dependencies and constructs a dependency graph to guide parallel commit selection, enforcing conflict-aware scheduling that avoids jointly updating strongly coupled positions, while DEMASK~\citep{ringel2026dependency} learns a lightweight predictor to estimate pairwise conditional influence and bound joint mismatch. Despite these advances, existing methods either rely on auxiliary models, learned predictors, or graph construction heuristics, and often require additional computation beyond a single forward pass.

Structured prediction frameworks, including energy-based models and graphical models, provide a principled way to encode dependencies via pairwise or higher-order interactions~\citep{lecun2006tutorial}. In particular, Markov Random Fields and mean-field variational inference~\citep{wainwright2008graphical, jordan1999introduction} have been widely used to approximate joint configurations through iterative updates combining unary and pairwise terms. Similar formulations have been applied in vision and sequence modeling for enforcing consistency constraints, but typically require learned energy functions or additional parameterization.
Our method differs from prior work by introducing a principled decoding formulation based on pairwise interactions defined in the model’s output space. Combined with a variational mean-field procedure, it enables coordinated parallel token commitment without additional training or auxiliary models.
Distributional divergences such as KL and JSD are widely used to compare predictive distributions in neural models, including uncertainty estimation, calibration, and distillation. In our setting, JSD is not used as a training objective, instead, it serves as a bounded, inference-time proxy for overlap between masked-token predictive distributions.

\section{Background: Discrete Diffusion Language Models}
\label{subsec:background}

Discrete Diffusion Language Models (dLLMs) define a generative process over a sequence of $L$ tokens $x = (x_1, \dots, x_L) \in \mathcal{V}^L$, where $\mathcal{V}$ is a discrete vocabulary, by reversing a discrete-state corruption process. Let $x_0$ be the clean data and $x_T$ be a fully corrupted sequence, typically where all tokens are replaced by a special \mask token. The forward process $q(x_t \mid x_{t-1})$ gradually adds noise, while the model $\Phi_\theta$ learns to approximate the reverse transition $p_\theta(x_{t-1} \mid x_t)$.

\paragraph{Iterative Denoising.} At any denoising step $t \in \{T, \dots, 1\}$, let $\mathcal{U}_t = \{i \in \{1, \dots, L\} : x_{t,i} = \mask\}$ denote the set of indices that remain masked. The model $\Phi_\theta$ predicts a categorical distribution over the vocabulary $\mathcal{V}$ for each masked position $i \in \mathcal{U}_t$:
\begin{equation}
    p_\theta(x_{i} \mid x_t) = \text{Cat}(x_i; \pi_{t,i}), \quad \forall i \in \mathcal{U}_t,
\end{equation}
where $\pi_{t,i} \in \Delta^{|\mathcal{V}|-1}$ is the predicted probability vector. In standard iterative decoding, a subset of positions $\mathcal{C}_t \subseteq \mathcal{U}_t$, referred to as the \emph{commit set}, is selected to be unmasked. The tokens at these positions are realized and fixed for the subsequent step $t-1$.

\paragraph{The Parallelism-Consistency Trade-off.} The efficiency of dLLMs stems from the ability to commit multiple tokens in parallel ($|\mathcal{C}_t| > 1$). However, most dLLMs employ a factorized reverse policy:
\begin{equation}
    p_\theta(x_{\mathcal{C}_t} \mid x_t) \approx \prod_{i \in \mathcal{C}_t} p_\theta(x_i \mid x_t).
\end{equation}
This independence assumption holds only when the selected positions in $\mathcal{C}_t$ do not interact under the current context $x_{\setminus \mathcal{U}_t}$. In practice, tokens often exhibit strong interactions arising from syntactic and semantic constraints.
As a result, aggressive parallelism can produce \emph{joint inconsistency}~\citep{ringel2026dependency,dawn}, where tokens that are individually probable form incompatible configurations when committed simultaneously.
The failure mode of aggressive parallel decoding is that adjacent or syntactically related masked positions often share similar predictive distributions; committing them simultaneously from independent marginals produces redundant or mutually incompatible tokens. JSD between predictive distributions provides a direct, low-cost test for this overlap, computed entirely from the single forward pass already performed by the model.
\section{Methodology}
\label{sec:methodology}
\subsection{Pairwise Interaction Commit Selection}

\label{subsec:method}

We treat the selection of the commit set $\mathcal{C}_t$ as a structured inference problem. We define a vector of binary indicators $S = (s_1, \dots, s_m) \in \{0, 1\}^m$, where $m = |\mathcal{U}_t|$ is the number of currently masked positions. Here, $s_i = 1$ indicates that position $i$ is selected for the commit set $\mathcal{C}_t$, and $s_i = 0$ otherwise.

\subsubsection{Energy-Based Formulation}
We define a pairwise Markov Random Field (MRF)~\citep{wainwright2008graphical}  over the commit indicators $S$ to capture the trade-off between local confidence and joint compatibility. The probability of a commit configuration $S$ is given by the Gibbs distribution:
\begin{equation}
    P(S; \mathcal{U}_t) = \frac{1}{Z} \exp \left(
    \sum_{s_i \in S} \phi_i(s_i)
    + \sum_{\substack{s_i,s_j \in S \\ i < j}} \phi_{ij}(s_i, s_j)
    \right).
\end{equation}
where $Z$ is the partition function, $\phi_i$ is the unary potential, and $\phi_{ij}$ is the pairwise potential.

\paragraph{Unary Potentials (Local Confidence).} The unary potential $\phi_i(s_i) = c_i s_i$ encourages committing positions with high local certainty. We define the local confidence $c_i$ as the log-margin between the top-2 predicted tokens:
\begin{equation}
    c_i = \log \pi_i(v_i^{(1)}) - \log \pi_i(v_i^{(2)}),
\end{equation}
where $v_i^{(1)}$ and $v_i^{(2)}$ denote the most likely and second most likely tokens under the predictive distribution $\pi_i$, respectively.
where $v_i^{(1)}$ and $v_i^{(2)}$ are the indices of the most likely and second most likely tokens at position $i$, respectively.
\begin{algorithm}[t]
\caption{Mean-Field Parallel Decoding}
\label{alg:ours}
\KwIn{Masked indices $\mathcal{U}_t$, logits $\{\ell_i\}_{i \in \mathcal{U}_t}$, parameters $ \tau, R$}
\KwOut{Commit set $\mathcal{C}_t$}
\tcp{1. Local Confidence Estimation}
$\pi_{t,i} \leftarrow \text{softmax}(\ell_i)$ for all $i \in \mathcal{U}_t$\;
$c_i \leftarrow \log \pi_{t,i}(v_i^{(1)}) - \log \pi_{t,i}(v_i^{(2)})$\;
\tcp{2. Pairwise Compatibility Construction}
$\tilde{D}_{ij} \gets (1 - \text{JSD}(\pi_{t,i}, \pi_{t,j})/\ln 2)$ for $i \neq j$\;
$D \gets \tilde{D} / \max(\tilde{D})$\;
\tcp{3. Mean-Field Relaxation}
$q^{(0)} \gets \sigma( c)$\;
\For{$r = 0$ \KwTo $R-1$}{
    $q^{(r+1)} \gets \sigma(c - D q^{(r)})$\;
}
\tcp{4. Thresholding}
$\mathcal{C}_t \gets \{i \in \mathcal{U}_t : q_i^{(R)} \ge \tau\}$\;
\Return{$\mathcal{C}_t$}\;
\end{algorithm}

\paragraph{Pairwise Potentials (Predictive Overlap).} The pairwise potential $\phi_{ij}(s_i, s_j) = - D_{ij} s_i s_j$, where $s_i,s_j \in \{0,1\}$ denote the commit indicators for masked positions $i$ and $j$, penalizes the simultaneous commitment of strongly coupled positions. We construct the pairwise matrix $D \in [0, 1]^{m \times m}$ using the normalized Jensen-Shannon Divergence (JSD) between the predicted distributions $\pi_{t,i}$ and $\pi_{t,j}$:
\begin{equation}
    \tilde{D}_{ij} = 1 - \frac{\mathrm{JSD}(\pi_{t,i}, \pi_{t,j})}{\ln 2},
\end{equation}
where the JSD is defined as:
\begin{equation}
    \text{JSD}(p, q) = \frac{1}{2}\text{KL}(p \| \frac{p+q}{2}) + \frac{1}{2}\text{KL}(q \| \frac{p+q}{2}).
\end{equation}
We set $D_{ij} = \tilde{D}_{ij} / \max_{a,b} \tilde{D}_{ab}$ for $i \neq j$ to normalize the scores within the current active set, and $D_{ii} = 0$ to avoid self-inhibition. This ``antiferromagnetic'' coupling ensures that if two positions have similar output distributions (high $D_{ij}$), they compete for commitment, preventing redundant or inconsistent parallel updates.

\subsubsection{Variational Mean-Field Inference}
Since finding the MAP assignment $S^* = \arg\max P(S)$ is NP-hard for general $D$~\citep{wainwright2008graphical}, we employ a mean-field approximation. We approximate $P(S)$ with a fully factorized distribution $q(S) = \prod_{i=1}^m q_i^{s_i}(1-q_i)^{1-s_i}$, where $q_i \in [0, 1]$ represents the \emph{commit intensity} (the marginal probability that $s_i=1$). We maximize the Evidence Lower Bound (ELBO):
\begin{equation}
    \mathcal{L}(q) = \mathbb{E}_q \left[ \sum_i  c_i s_i - \sum_{i < j}  D_{ij} s_i s_j \right] + \mathcal{H}(q),
\end{equation}
where $\mathcal{H}(q) = \sum_i H(q_i)$ is the sum of binary entropies $H(q_i) = -q_i \log q_i - (1-q_i) \log(1-q_i)$. Expanding the expectation:
\begin{equation}
    \mathcal{L}(q) = \sum_{i=1}^m  c_i q_i - \frac{1}{2} \sum_{i \neq j} D_{ij} q_i q_j + \sum_{i=1}^m \left[ -q_i \log q_i - (1-q_i) \log(1-q_i) \right].
\end{equation}
Setting the partial derivative $\frac{\partial \mathcal{L}}{\partial q_i} = 0$ yields the fixed-point equation:
\begin{equation}
     c_i - \sum_{j \neq i} D_{ij} q_j - \log \frac{q_i}{1-q_i} = 0,
\end{equation}
which implies:
\begin{equation}
    q_i = \sigma \left(c_i -  \sum_{j \neq i} D_{ij} q_j \right),
\end{equation}
where $\sigma(x) = (1+e^{-x})^{-1}$ is the logistic sigmoid function.
The commit set $\mathcal{C}_t$ is obtained by iterating the mean-field update and thresholding the final intensities. The complete procedure is summarized in Algorithm~\ref{alg:ours}.

\paragraph{Complexity and Runtime Overhead.} 
The proposed framework is entirely training-free and operates within the scope of a single forward pass of the base discrete diffusion model, requiring no auxiliary predictors or fine-tuning stages. The primary computational overhead arises from the $O(m^2 |\mathcal{V}|)$ complexity of the pairwise Jensen–Shannon Divergence (JSD) computation, where $m$ is the number of masked positions and $|\mathcal{V}|$ is the vocabulary size. However, this operation consists of independent, element-wise calculations that are highly amenable to parallelization on modern GPU accelerators, ensuring that the additional latency is negligible compared to the model's execution time. In practice, we observe that setting the iteration count to $R=2$ is sufficient for the mean-field intensity values to reach a stable fixed point. This rapid convergence ensures that the coordination of the commit set does not bottleneck the overall inference pipeline, maintaining the high-throughput advantages of parallel decoding while preserving more reliable parallel commitments.

\section{Experiments}
\label{sec:exps}
\begin{table*}[t]
\setlength{\tabcolsep}{2.pt}
  \centering
  \small
  \caption{Performance comparison between Mean-Field Parallel Decoding and baselines across GSM8K, MATH, HumanEval, and MBPP datasets and \texttt{LLaDA-8B-Instruct}, \texttt{LLaDA-1.5}, and \texttt{Dream-v0-Instruct-7B} models. We report Accuracy, TPS, and Speedup.}
  \label{tab:main_results}

    \begin{tabular}{c*{4}{ccc}}
      \toprule
        & \multicolumn{3}{c}{GSM8K}
        & \multicolumn{3}{c}{MATH}
        & \multicolumn{3}{c}{HumanEval}
        & \multicolumn{3}{c}{MBPP}\\
      \cmidrule(lr){2-4} \cmidrule(lr){5-7} \cmidrule(lr){8-10} \cmidrule(lr){11-13} 
      Method
        & Acc.$\uparrow$ & TPS$\uparrow$ & Speedup$\uparrow$
        & Acc.$\uparrow$ & TPS$\uparrow$ & Speedup$\uparrow$
        & Acc.$\uparrow$ & TPS$\uparrow$ & Speedup$\uparrow$
        & Acc.$\uparrow$ & TPS$\uparrow$ & Speedup$\uparrow$\\
      \midrule
      \multicolumn{13}{c}{LLaDA-8B-Instruct}\\
      \midrule
      Entropy   & 79.15 & 15.87 & 1.00$\times$ & 33.36 & 23.59 & 1.00$\times$ & 40.24 & 38.02 & 1.00$\times$ & 29.4  & 15.72 & 1.00$\times$\\
      KLASS     & 75.05  & 36.23 & 2.27$\times$ & 31.80  & 42.20 & 1.79$\times$ & 39.63  & 70.30 & 1.83$\times$ & 27.2  & 35.60 & 2.28$\times$\\
      LocalLeap & 77.71 & 68.96 & 4.32$\times$ & 32.96 & 77.87 & 3.30$\times$ & 40.24 & 152.62 & 4.04$\times$ & 30.6 & 72.64 & 4.60$\times$\\
      DAWN      & 77.78 & 69.76 & 4.38$\times$ & 32.34 & 80.40 & 3.41$\times$ & 40.24 & 152.46 & 4.05$\times$ & 28.0 & 74.46 & 4.87$\times$\\
      Ours      & 76.34 & \textbf{78.75} & \textbf{4.93$\times$} & 31.10 & \textbf{96.37} & \textbf{4.07$\times$} & 40.70 & \textbf{180.81} & \textbf{4.88$\times$} & 28.0 & \textbf{81.11} & \textbf{6.10$\times$}\\
      \midrule
      \multicolumn{13}{c}{LLaDA-1.5}\\
      \midrule
      Entropy   & 81.80 & 14.90 & 1.00$\times$ & 33.62 & 20.91 & 1.00$\times$ & 44.51 & 13.04 & 1.00$\times$ & 39.2 & 5.77  & 1.00$\times$\\
      KLASS     & 79.37 & 35.22 & 2.37$\times$ & 32.50 & 38.52 & 1.87$\times$ & 43.90 & 18.94 & 1.48$\times$ & 29.8 & 18.25 & 4.08$\times$\\
      LocalLeap & 80.36 & 65.69 & 4.41$\times$ & 33.02 & 68.87 & 3.29$\times$ & 42.07 & 41.91 & 3.23$\times$ & 38.0 & 43.50 & 7.57$\times$\\
      DAWN      & 79.60 & 66.74 & 4.47$\times$ & 32.16 & 70.90 & 3.39$\times$ & 42.68 & 41.50 & 3.20$\times$ & 37.4 & 45.36 & 7.90$\times$\\
      Ours      & 80.74 & \textbf{75.18} & \textbf{4.96$\times$} & 32.46 & \textbf{86.71} & \textbf{4.10$\times$} & 42.24 & \textbf{54.91} & \textbf{4.32$\times$} & 38.6 & \textbf{50.95} & \textbf{8.62$\times$}\\
      \midrule 
      \multicolumn{13}{c}{Dream-v0-Instruct-7B}\\
      \midrule
      Entropy   & 75.51 & 11.15 & 1.00$\times$ & 38.30 & 28.18 & 1.00$\times$ & 53.04 & 32.05 & 1.00$\times$ & 54.20 & 13.40 & 1.00$\times$\\
      KLASS     & 73.54 & 19.61 & 1.82$\times$ & 36.64 & 29.26 & 1.70$\times$ & 53.65 & 38.63 & 1.83$\times$ & 56.60 & 29.00 & 1.30$\times$\\
      LocalLeap & 73.61 & 49.80 & 4.49$\times$ & 38.38 & 68.62 & 2.51$\times$ & 56.09 & 84.42 & 3.05$\times$ & 55.00 & 66.86 & 5.39$\times$\\
      DAWN      & 73.31 & 50.15 & 4.53$\times$ & 38.42 & 69.77 & 2.57$\times$ & 54.87 & 89.56 & 3.26$\times$ & 55.60 & 69.50 & 5.62$\times$\\
      Ours      & 74.90 & \textbf{54.75} & \textbf{4.91$\times$} & 40.44 & \textbf{74.38} & \textbf{3.61$\times$} & \_\_  & \_\_  & \_\_  & 54.40 & \textbf{75.68} & \textbf{5.80$\times$}\\
      \bottomrule
    \end{tabular}
\end{table*}
We empirically evaluate the proposed  decoding framework under standard discrete diffusion generation settings. Our experiments focus on quantifying the impact of structured commit selection on the quality-latency trade-off, compared to marginal-based and heuristic dependency-aware baselines.

\subsection{Setups}
\paragraph{Datasets.} We evaluate the proposed decoding method on four standard benchmarks spanning mathematical reasoning and code generation. For reasoning, we use GSM8K~\citep{cobbe2021training} and MATH~\citep{hendrycks2021measuring}, which test grade-school arithmetic and competition-level mathematical problem solving, respectively. For code generation, we use HumanEval~\citep{chen2021evaluating} and MBPP~\citep{austin2021program}, which evaluate functional correctness on Python programming tasks. Following prior work on diffusion language model decoding~\citep{dawn}, we report results across multiple model variants, including both base and instruction-tuned models when available. All decoding methods are evaluated under the same model, dataset split, generation budget, and stopping criteria to ensure that observed differences reflect the decoding strategy rather than differences in evaluation protocol.
\paragraph{Models.}
We evaluate all decoding methods on three recent dLLM backbones: \drean\footnote{\url{https://huggingface.co/Dream-org/Dream-v0-Instruct-7B}}~\citep{ye2025dream}, \llada\footnote{\url{https://huggingface.co/GSAI-ML/LLaDA-1.5}}~\citep{zhu2025llada}, and \lladanew\footnote{\url{https://huggingface.co/GSAI-ML/LLaDA-8B-Instruct}}~\citep{zhu2025llada}. These models cover both instruction-tuned and base settings, allowing us to assess whether the proposed commit-selection rule generalizes across different dLLM families and model scales. For each backbone, we use the publicly released checkpoint and keep the model weights fixed, only the decoding strategy is varied across methods.
\paragraph{Baselines.}
We compare against representative decoding strategies that cover uncertainty-based decoding, confidence-based acceleration, locality-aware scheduling, and dependency-aware parallel commitment. The Entropy baseline commits tokens according to position-wise predictive uncertainty and serves as the reference method for computing relative speedup. KLASS~\citep{kim2026klass} selects tokens using a KL-based confidence criterion, providing a stronger marginal-selection baseline. LocalLeap~\citep{kong2025accelerating} uses a training-free, anchor-guided schedule: it identifies high-confidence tokens as anchors and relaxes the commit criterion for nearby positions within a bounded local neighborhood, thereby exploiting local determinism without explicitly constructing a pairwise dependency matrix. DAWN~\citep{dawn} introduces dependency-aware scheduling heuristics designed to increase safe parallelism during diffusion decoding. Together, these baselines allow us to compare the proposed structured mean-field commit rule against marginal confidence selection, local confidence propagation, and existing dependency-aware parallel decoding methods. We do not include DEMASK~\citep{ringel2026dependency} in the main comparison because it introduces a trainable dependency predictor, while our focus is on training-free decoding methods that operate directly on fixed pretrained diffusion language models.

\paragraph{Metrics.}
We report three metrics: task accuracy, tokens per second (TPS), and speedup relative to the Entropy baseline. Accuracy measures task-level generation quality, using the standard evaluation protocol for each benchmark. TPS measures decoding throughput and is computed as the number of generated tokens divided by wall-clock decoding time. Speedup is computed as the ratio between the TPS of a given method and the TPS of Entropy under the same model and benchmark. Reporting these metrics jointly characterizes the quality--latency trade-off: a desirable decoder should improve throughput without causing a disproportionate degradation in task accuracy.
\paragraph{Hardware and Implementation Details.}
All experiments are implemented within the DAWN\footnote{\url{https://github.com/lizhuo-luo/DAWN}} diffusion decoding codebase. For all baselines and our method, we use identical model weights, tokenizer settings, decoding budgets, stopping criteria, and block-sampling protocol. Throughput is measured end-to-end during decoding, including the overhead of commit selection and mean-field inference. For LLaDA models, we tune the decoding parameters on 100 samples from the GSM8K training set and use block size $20$, commit threshold $\tau=0.85$, and $R=2$ mean-field iterations. All experiments are conducted on a cluster with $8\times$ NVIDIA H100 GPUs.

\begin{table}[t]
\centering
\small
\caption{
Ablation of the proposed decoding components and inference controls using Dream-7B-Instruct on GSM8K
($\tau{=}0.9$, $R{=}2$, Block size = 32).
We report task accuracy, throughput, and the number of denoising function evaluations (NFE).
}
\label{tab:ablation_main}
\begin{tabular}{lcccc}
\toprule
Variant 
& Acc. $\uparrow$ 
& TPS $\uparrow$ 
& NFE $\downarrow$ \\
\midrule
Baseline (Entropy) 
& 75.51 & 11.15 & 256.0 \\
\midrule
No pairwise interaction 
& 72.40 & 56.20 & 48.7 \\
Uniform interaction matrix 
& 71.95 & 51.92 & 52.6 \\
One-shot update, $R{=}1$ 
& 54.51 & 20.11 & 140.0 \\
Ours 
& 73.24 & 49.13 & 54.3 \\
\bottomrule
\end{tabular}
\end{table}
\subsection{Quantitative Results} Table~\ref{tab:main_results} summarizes the quality and latency trade-off across the reasoning and code-generation benchmarks. The main trend is that the proposed method improves throughput in every evaluation setting while keeping task accuracy close to the base. Across all of the entries, our method obtains the highest TPS among all reported methods. The average speedup relative to entropy-based decoding is $5.12\times$, with a range from $3.61\times$ on \texttt{Dream-v0-Instruct-7B} for MATH to $8.62\times$ on \texttt{LLaDA-1.5} for MBPP. These results indicate that the proposed Mean-Field Parallel Decoding rule increases parallelism substantially beyond marginal confidence selection.

The comparison with Entropy and KLASS highlights the benefit of moving beyond position-wise commit criteria. Entropy is generally the most conservative baseline: it often preserves strong accuracy, but does so with substantially lower throughput. KLASS improves efficiency by using a stronger confidence-based selection rule, yet its gains remain below those of methods that incorporate dependency or scheduling information. In contrast, the proposed method consistently shifts the trade-off toward higher throughput while keeping accuracy within a narrow margin of the entropy baseline. In some cases, it also matches or exceeds Entropy in accuracy, suggesting that additional parallelism does not necessarily compromise generation quality when commit decisions account for token interactions.

Compared with dependency-aware baselines such as LocalLeap and DAWN, the proposed method continues to provide a consistent throughput advantage. This is notable because these baselines already introduce stronger scheduling or dependency heuristics than purely marginal selection. The improvement suggests that the mean-field commit procedure selects more efficient parallel commit sets while avoiding the need for an auxiliary verifier or additional model forward passes. Overall, the results support the view that structured commit selection is an effective mechanism for improving the efficiency of diffusion decoding while preserving competitive task accuracy.
\subsection{Ablation Studies}
We conduct a set of ablation studies to evaluate the contribution of each component in the proposed decoding method. As shown in Table~\ref{tab:ablation_main}, the entropy baseline achieves the highest accuracy but requires substantially more denoising function evaluations, reflecting its conservative token-commitment behavior. Removing the pairwise interaction term greatly improves throughput and reduces NFE, but it also leads to a larger accuracy drop, indicating that local confidence alone is not sufficient to preserve generation quality under aggressive parallel decoding.

The uniform interaction matrix provides a further control for the pairwise interaction term. Its weaker performance compared with the full method suggests that the structure of the JSD-derived interaction matrix is critical, rather than merely the presence of a generic pairwise penalty. The one-shot update variant with $R{=}1$ performs substantially worse, showing that a single mean-field update is insufficient to produce a reliable commit set. Overall, the full method provides the best balance among accuracy, throughput, and NFE, supporting the use of structured pairwise interactions together with iterative mean-field refinement.

\begin{figure}[t]
\centering
\begin{minipage}[t]{0.48\textwidth}
    \centering
    \label{fig:your_figure1}
    \includegraphics[width=0.8\textwidth, height=3cm]{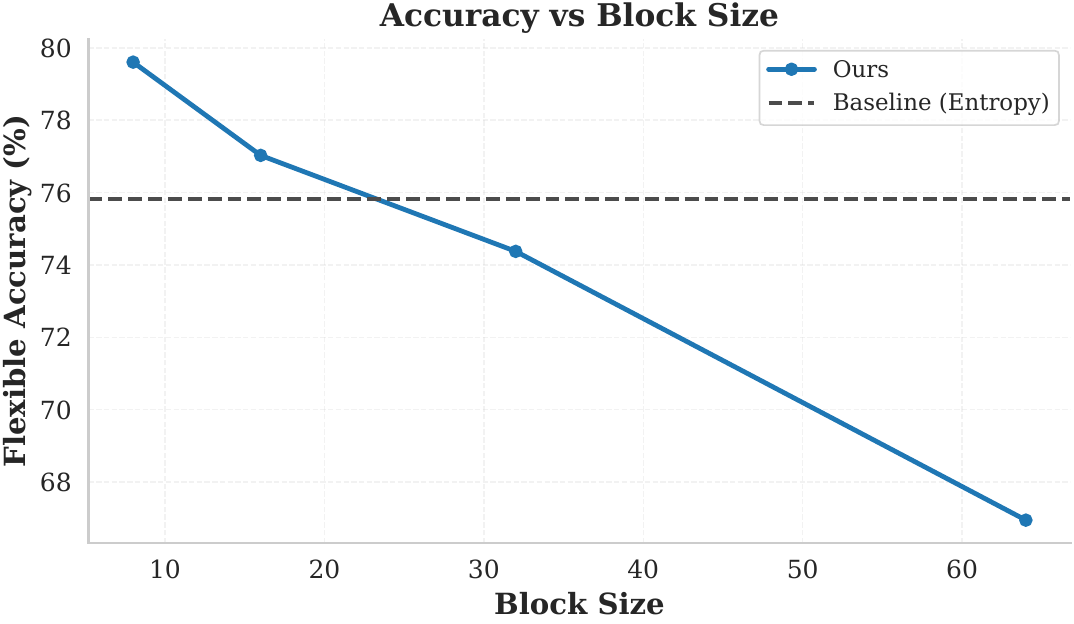}
    
    \vspace{0.2cm}
    
    \includegraphics[width=0.8\textwidth, height=3cm]{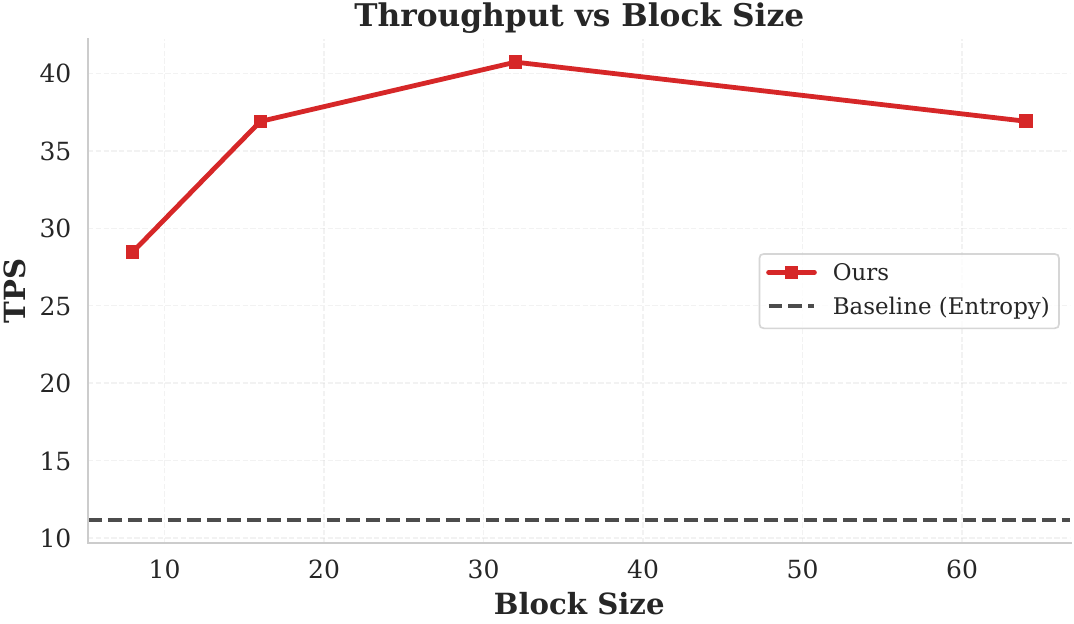}
    \caption{Block-size sensitivity on GSM8K using \texttt{Dream-7B-Instruct}. We evaluate our method with $R=2$ and $\tau=0.95$.}
    \label{fig:blockablation}

\end{minipage}
\hfill
\begin{minipage}[t]{0.48\textwidth}
\captionof{table}{
Sensitivity to decoding controls on GSM8K using \drean. We vary the commit threshold $\tau$ and mean-field iterations $R$ around the default setting $\tau=0.95$, $R=2$, and block size $20$.}
    \label{tab:sensitivity}
    
    \vspace{0.3cm}

    \begin{tabular}{@{}l@{~~~}r@{~~~}c@{~~~}c@{~~~}c@{~~~}c@{}}
    \toprule
    \textbf{Setting} & \textbf{Value} & \textbf{Acc.} & \textbf{TPS} & \textbf{Speed} & \textbf{NFE} \\
    \midrule
    Baseline & -- & 75.51 & 11.15 & 1.00× & 256.0 \\
    \midrule
    \multirow{5}{*}{$\tau$} 
    & 0.75 & 66.41 & 57.78 & 5.23× & 47.8 \\
    & 0.80 & 71.42 & 52.46 & 4.75× & 52.5 \\
    & 0.85 & 72.86 & 47.57 & 4.35× & 57.5 \\
    & 0.90 & 75.36 & 42.84 & 3.90× & 64.1 \\
    & 0.95 & 77.03 & 36.89 & 3.38× & 74.1 \\
    \midrule
    \multirow{4}{*}{$R$} 
    & 2 & 77.03 & 36.88 & 3.38× & 74.1 \\
    & 4 & 75.36 & 35.64 & 3.28× & 76.2 \\
    & 6 & 75.06 & 34.96 & 3.23× & 77.3 \\
    & 8 & 74.53 & 34.70 & 3.20× & 78.2 \\
    \bottomrule
    \end{tabular}
\end{minipage}
\end{figure}

Furthermore, we analyze the sensitivity of the proposed decoder to its main inference controls on GSM8K, as shown in Table~\ref{tab:sensitivity}. The commit threshold $\tau$ directly controls the aggressiveness of parallel token commitment. Lower thresholds produce higher throughput and fewer denoising steps, but they also reduce accuracy, indicating that overly aggressive commitment can amplify joint inconsistency. Increasing $\tau$ makes the decoder more conservative, improving accuracy while gradually reducing TPS and speedup.
We also vary the number of mean-field iterations $R$. The results show that using $R{=}2$ is sufficient in this setting; additional iterations provide little benefit and slightly reduce throughput due to the added inference cost. Together with the threshold analysis, this shows that the decoder exposes a smooth and interpretable quality--atency trade-off: more conservative settings preserve accuracy, while more aggressive settings reduce NFE and improve throughput. Overall, the sensitivity results indicate that the method behaves predictably across hyperparameter settings and does not depend on a narrow configuration to obtain favorable quality-latency trade-offs.

Finally, Figure~\ref{fig:blockablation} analyzes the effect of block size on \drean\ using GSM8K, with $R=2$ and $\tau=0.95$ fixed. Increasing the block size improves throughput by expanding the set of positions considered for parallel commitment at each denoising step. Accuracy remains competitive across moderate block sizes, while very large blocks reflect the expected trade-off of more aggressive decoding. The intermediate settings form the Pareto-optimal region, achieving substantially higher throughput than entropy decoding while preserving strong reasoning performance. This indicates that the proposed mean-field commit rule provides an effective operating regime for scaling parallelism in discrete diffusion decoding.

\section{Limitations}
\label{sec:lim}
Mean-Field Parallel Decoding mitigates joint inconsistency by treating token commitment as a structured selection problem rather than as independent marginal decisions. The method is training-free and relies only on the model's predictive distributions, but it still has several limitations. First, constructing the pairwise interaction matrix costs $O(m^2|\mathcal{V}|)$ per denoising step. In block-based decoding, this cost is controlled because the active set $m$ is bounded by the block size, making the computation practical and parallelizable on modern GPUs. For substantially larger blocks, approximate distributional comparisons may be needed. Another limitation is that the JSD-based interaction score is a lightweight proxy for unsafe simultaneous commitment; it captures pairwise predictive overlap but does not explicitly model higher-order structure or recover the true joint conditional distribution. Finally, the method retains a quality-latency trade-off through the block size and commit threshold. Our experiments show favorable Pareto-optimal operating points, but the best configuration may vary across models, tasks, and decoding budgets.
\section{Conclusions}

In this work, we introduced Mean-Field Parallel Decoding, a training-free framework that mitigates the joint inconsistency bottleneck in discrete diffusion language models. By formulating parallel token selection as a variational mean-field optimization, we replace independent marginal updates with a structured process that balances local confidence against global compatibility. This approach enables the decoder to safely unmask a larger number of tokens in each step by using the model’s own predictive distributions to suppress conflicting updates. Our experiments demonstrate that this coordinated selection significantly improves the quality-latency trade-off across reasoning and coding tasks without requiring auxiliary models or retraining. Ultimately, Mean-Field Parallel Decoding provides a principled, single-pass mechanism for improving the throughput of diffusion-based generation while preserving the linguistic coherence essential for complex text synthesis.

\vspace{-4pt}
\section{Acknowledgments}
\vspace{-2pt}
This work was supported by a grant from the Tel Aviv University Center for AI and Data Science (TAD). This research was also supported by the Ministry of Innovation, Science \& Technology ,Israel (1001576154) and the Michael J. Fox
Foundation (MJFF-022407). The contribution of Tamim Zoabi is part of a PhD thesis research conducted at Tel Aviv University.







\appendix

\bibliographystyle{plainnat} 
\bibliography{neurips_2026}

\end{document}